\documentclass[11pt,a4paper]{article}
\usepackage[hyperref]{acl2019}
\usepackage{times}
\usepackage{latexsym}
\usepackage{amsmath}
\usepackage{graphicx}
\usepackage{filecontents}
\usepackage[labelfont=bf]{caption}
\usepackage{pgfplots,tikz}
\usepackage{hyperref}
\usepgfplotslibrary{colorbrewer}
\pgfplotsset{
	cycle list/Dark2,
	cycle multiindex* list={
		mark list*\nextlist
		Dark2\nextlist
	},
}
\pgfplotsset{compat=1.14}
\usepackage{etoolbox}
\usepackage{makecell}
\usepackage{siunitx}
\usepackage{paralist}
\usepackage{url}
\usepackage{amsfonts}
\usepackage{booktabs}
\usepackage{multirow}
\usepackage{subcaption}
\usepackage{tikz}
\usepackage{xspace,mfirstuc,tabulary}
\usetikzlibrary{bayesnet}

\allowdisplaybreaks

\newcommand{\secref}[1]{\S\ref{#1}\xspace}
\newcommand{\tabref}[2][]{Table#1~\ref{#2}\xspace}
\newcommand{\figref}[2][]{Figure#1~\ref{#2}\xspace}

\newcommand*\rot{\rotatebox{90}}

\newcommand{\method}[1]{\texttt{#1}\xspace}

\newcommand{\fasttext}{\method{fastText}}
\newcommand{\muse}{\method{MUSE}}

\newcommand{\fone}{F$_1$\xspace}

\newcommand{\bwet}{\method{BWET}}
\newcommand{\wikian}{\method{Wikiann}}
\newcommand{\lowsup}{\method{LSup}}
\newcommand{\highsup}{\method{HSup}}
\newcommand{\rare}{\method{RaRe}}
\newcommand{\raress}[1]{\method{RaRe}$^{#1}$}
\newcommand{\sbest}{\method{Oracle}}
\newcommand{\tok}{\method{tok}}
\newcommand{\ent}{\method{ent}}
\newcommand{\mv}{\method{MV}}
\newcommand{\rareuns}{\rare$_{\texttt{uns}}$\xspace}
\newcommand{\rareunsss}[1]{\rare$_{\texttt{uns}}^{#1}$\xspace}
\newcommand{\mvtok}[1][]{$\mv^\tok_{#1}$\xspace}
\newcommand{\mvent}[1][]{$\mv^\ent_{#1}$\xspace}
\newcommand{\bea}{\method{BEA}}
\newcommand{\beauns}[1][]{$\bea^{#1}_{\method{uns}}$\xspace}
\newcommand{\beaunstwice}[1][]{$\bea^{#1}_{\method{uns}\times2}$\xspace}
\newcommand{\beasup}[1][]{$\bea^{#1}_{\method{sup}}$\xspace}
\newcommand{\beaunsent}[1][]{$\bea^\ent_{\method{uns}#1}$\xspace}
\newcommand{\beasupent}[1][]{$\bea^\ent_{\method{sup}#1}$\xspace}

\newcommand{\sagg}{\rare}

\newcommand{\ssagg}{\hyperref[sec:sagg]{\sagg}\xspace}

\newcommand{\rsagg}{\hyperref[tab:results]{\sagg}\xspace}
\newcommand{\rbeasup}{\hyperref[tab:results]{\beasup}\xspace}

\newcommand{\zaf}{\href{https://www.loc.gov/standards/iso639-2/php/langcodes_name.php?iso_639_1=af&lang=Afrikaans}{af}\xspace}
\newcommand{\zar}{\href{https://www.loc.gov/standards/iso639-2/php/langcodes_name.php?iso_639_1=ar&lang=Arabic}{ar}\xspace}
\newcommand{\zbg}{\href{https://www.loc.gov/standards/iso639-2/php/langcodes_name.php?iso_639_1=bg&lang=Bulgarian}{bg}\xspace}
\newcommand{\zbn}{\href{https://www.loc.gov/standards/iso639-2/php/langcodes_name.php?iso_639_1=bn&lang=Bengali}{bn}\xspace}
\newcommand{\zbs}{\href{https://www.loc.gov/standards/iso639-2/php/langcodes_name.php?iso_639_1=bs&lang=Bosnian}{bs}\xspace}
\newcommand{\zca}{\href{https://www.loc.gov/standards/iso639-2/php/langcodes_name.php?iso_639_1=ca&lang=Catalan}{ca}\xspace}
\newcommand{\zcs}{\href{https://www.loc.gov/standards/iso639-2/php/langcodes_name.php?iso_639_1=cs&lang=Czech}{cs}\xspace}
\newcommand{\zda}{\href{https://www.loc.gov/standards/iso639-2/php/langcodes_name.php?iso_639_1=da&lang=Danish}{da}\xspace}
\newcommand{\zde}{\href{https://www.loc.gov/standards/iso639-2/php/langcodes_name.php?iso_639_1=de&lang=German}{de}\xspace}
\newcommand{\zel}{\href{https://www.loc.gov/standards/iso639-2/php/langcodes_name.php?iso_639_1=el&lang=Greek}{el}\xspace}
\newcommand{\zen}{\href{https://www.loc.gov/standards/iso639-2/php/langcodes_name.php?iso_639_1=en&lang=English}{en}\xspace}
\newcommand{\zes}{\href{https://www.loc.gov/standards/iso639-2/php/langcodes_name.php?iso_639_1=es&lang=Spanish}{es}\xspace}
\newcommand{\zet}{\href{https://www.loc.gov/standards/iso639-2/php/langcodes_name.php?iso_639_1=et&lang=Estonian}{et}\xspace}
\newcommand{\zfa}{\href{https://www.loc.gov/standards/iso639-2/php/langcodes_name.php?iso_639_1=fa&lang=Persian}{fa}\xspace}
\newcommand{\zfi}{\href{https://www.loc.gov/standards/iso639-2/php/langcodes_name.php?iso_639_1=fi&lang=Finnish}{fi}\xspace}
\newcommand{\zfr}{\href{https://www.loc.gov/standards/iso639-2/php/langcodes_name.php?iso_639_1=fr&lang=French}{fr}\xspace}
\newcommand{\zhe}{\href{https://www.loc.gov/standards/iso639-2/php/langcodes_name.php?iso_639_1=he&lang=Hebrew}{he}\xspace}
\newcommand{\zhi}{\href{https://www.loc.gov/standards/iso639-2/php/langcodes_name.php?iso_639_1=hi&lang=Hindi}{hi}\xspace}
\newcommand{\zhr}{\href{https://www.loc.gov/standards/iso639-2/php/langcodes_name.php?iso_639_1=hr&lang=Croatian}{hr}\xspace}
\newcommand{\zhu}{\href{https://www.loc.gov/standards/iso639-2/php/langcodes_name.php?iso_639_1=hu&lang=Hungarian}{hu}\xspace}
\newcommand{\zid}{\href{https://www.loc.gov/standards/iso639-2/php/langcodes_name.php?iso_639_1=id&lang=Indonesian}{id}\xspace}
\newcommand{\zit}{\href{https://www.loc.gov/standards/iso639-2/php/langcodes_name.php?iso_639_1=it&lang=Italian}{it}\xspace}
\newcommand{\zlt}{\href{https://www.loc.gov/standards/iso639-2/php/langcodes_name.php?iso_639_1=lt&lang=Lithuanian}{lt}\xspace}
\newcommand{\zlv}{\href{https://www.loc.gov/standards/iso639-2/php/langcodes_name.php?iso_639_1=lv&lang=Latvian}{lv}\xspace}
\newcommand{\zmk}{\href{https://www.loc.gov/standards/iso639-2/php/langcodes_name.php?iso_639_1=mk&lang=Macedonian}{mk}\xspace}
\newcommand{\zms}{\href{https://www.loc.gov/standards/iso639-2/php/langcodes_name.php?iso_639_1=ms&lang=Malay}{ms}\xspace}
\newcommand{\znl}{\href{https://www.loc.gov/standards/iso639-2/php/langcodes_name.php?iso_639_1=nl&lang=Dutch}{nl}\xspace}
\newcommand{\zno}{\href{https://www.loc.gov/standards/iso639-2/php/langcodes_name.php?iso_639_1=no&lang=Norwegian}{no}\xspace}
\newcommand{\zpl}{\href{https://www.loc.gov/standards/iso639-2/php/langcodes_name.php?iso_639_1=pl&lang=Polish}{pl}\xspace}
\newcommand{\zpt}{\href{https://www.loc.gov/standards/iso639-2/php/langcodes_name.php?iso_639_1=pt&lang=Portuguese}{pt}\xspace}
\newcommand{\zro}{\href{https://www.loc.gov/standards/iso639-2/php/langcodes_name.php?iso_639_1=ro&lang=Romanian}{ro}\xspace}
\newcommand{\zru}{\href{https://www.loc.gov/standards/iso639-2/php/langcodes_name.php?iso_639_1=ru&lang=Russian}{ru}\xspace}
\newcommand{\zsk}{\href{https://www.loc.gov/standards/iso639-2/php/langcodes_name.php?iso_639_1=sk&lang=Slovak}{sk}\xspace}
\newcommand{\zsl}{\href{https://www.loc.gov/standards/iso639-2/php/langcodes_name.php?iso_639_1=sl&lang=Slovenian}{sl}\xspace}
\newcommand{\zsq}{\href{https://www.loc.gov/standards/iso639-2/php/langcodes_name.php?iso_639_1=sq&lang=Albanian}{sq}\xspace}
\newcommand{\zsv}{\href{https://www.loc.gov/standards/iso639-2/php/langcodes_name.php?iso_639_1=sv&lang=Swedish}{sv}\xspace}
\newcommand{\zta}{\href{https://www.loc.gov/standards/iso639-2/php/langcodes_name.php?iso_639_1=ta&lang=Tamil}{ta}\xspace}
\newcommand{\ztl}{\href{https://www.loc.gov/standards/iso639-2/php/langcodes_name.php?iso_639_1=tl&lang=Tagalog}{tl}\xspace}
\newcommand{\ztr}{\href{https://www.loc.gov/standards/iso639-2/php/langcodes_name.php?iso_639_1=tr&lang=Turkish}{tr}\xspace}
\newcommand{\zuk}{\href{https://www.loc.gov/standards/iso639-2/php/langcodes_name.php?iso_639_1=uk&lang=Ukrainian}{uk}\xspace}
\newcommand{\zvi}{\href{https://www.loc.gov/standards/iso639-2/php/langcodes_name.php?iso_639_1=vi&lang=Vietnamese}{vi}\xspace}

\newrobustcmd*{\mysquare}[1]{\tikz{\filldraw[draw=#1,fill=#1] (0,0)
		rectangle (0.2cm,0.2cm);}}
\newrobustcmd*{\mycircle}[1]{\tikz{\filldraw[draw=#1,fill=#1] (0,0) circle [radius=0.1cm];}}
\newrobustcmd*{\mytriangle}[1]{\tikz{\filldraw[draw=#1,fill=#1] (0,0) --
		(0.2cm,0) -- (0.1cm,0.2cm);}}
\newrobustcmd*{\myheart}[1]{\tikz{\filldraw[draw=#1,fill=#1] (0,0) heart[radius=0.1cm];}}

\definecolor{bblue}{HTML}{4F81BD}
\definecolor{rred}{HTML}{C0504D}
\definecolor{ggreen}{HTML}{9BBB59}
\definecolor{ppurple}{HTML}{9F4C7C}

\newcommand{\plotfile}[1]{
	\pgfplotstableread{#1}{\table}
	\pgfplotstablegetcolsof{#1}
	\pgfmathtruncatemacro\numberofcols{\pgfplotsretval-1}
	\pgfplotsinvokeforeach{1,...,\numberofcols}{
		\pgfplotstablegetcolumnnamebyindex{##1}\of{\table}\to{\colname}
		\addplot table [y index=##1] {#1}; 
	}
}

\graphicspath{{./figs/}}

\aclfinalcopy 
\def\aclpaperid{1966} 

\title{Massively Multilingual Transfer for NER}

\author{Afshin Rahimi\thanks{~Both authors  contributed equally to this work.} \qquad Yuan Li\footnotemark[1] \qquad Trevor Cohn\\
	School of Computing and Information Systems\\
	The University of Melbourne\\
	{\tt yuanl4@student.unimelb.edu.au}\\
	{\tt \{rahimia,t.cohn\}@unimelb.edu.au}
}

\date{}
\begin{document}
\maketitle
\begin{abstract}

In cross-lingual transfer, NLP models over one or more source languages are applied to a low-resource target language.
While most prior work has used a single source model or a few carefully selected models, here we consider a ``massive'' setting with many such models. This setting raises the problem of poor transfer, particularly from distant languages.
We propose two techniques for modulating the transfer, suitable for  zero-shot or few-shot learning, respectively.
Evaluating on named entity recognition, we show that our techniques are much more effective than strong baselines, including standard ensembling, and our unsupervised method rivals oracle selection of the single best individual model.\footnote{The code and the datasets will be made available at  \url{https://github.com/afshinrahimi/mmner}.}





\end{abstract}

\section{Introduction}
Supervised learning remains king in natural language processing, with most tasks requiring large quantities of annotated corpora.
The majority of the world's 6,000+ languages however have limited or no annotated text, and therefore much of the progress in NLP has yet to be realised widely.
Cross-lingual transfer learning is a technique which can compensate for the dearth of data, by transferring knowledge from high- to low-resource languages,
which has typically taken the form of  annotation projection over parallel corpora or other multilingual resources~\cite{yarowsky2001inducing,hwa2005bootstrapping},
or making use of transferable representations, such as phonetic transcriptions~\cite{bharadwaj2016phono}, closely related languages~\cite{I17-2016} or bilingual dictionaries~\cite{mayhew2017cheap,xie2018neural}.

Most methods proposed for cross-lingual transfer
rely on a single source language, which limits the transferable knowledge to only one
source.
The target language might be similar to many source languages, on the grounds of the script, word order, loan words etc, and transfer would benefit from these diverse sources of information.
There are a few exceptions, which use transfer from several languages, ranging from multitask learning~\cite{long2015low,ammar2016many,moe2017transfer}, and annotation projection from several languages~\cite{tackstrom2012nudging,fang2016distant,plank2018distant}.
However, to the best of our knowledge, none of these approaches adequately account for the quality of transfer, but rather ``weight'' the contribution of each language uniformly.

In this paper, we propose a novel method for zero-shot multilingual transfer, inspired by research in truth inference in crowd-sourcing, a related problem, in which the `ground truth' must be inferred from the outputs of several unreliable annotators \cite{dawid1979maximum}.
In this problem, the best approaches estimate each model's reliability, and their patterns of mistakes \cite{kim2012bayesian}.
Our proposed model adapts these ideas to a multilingual transfer setting, whereby we learn the quality of transfer, and language-specific transfer errors, in order to infer the best labelling in the target language, as part of a Bayesian graphical model.
The key insight is that while the majority of poor models make lots of mistakes, these mistakes are diverse, while the few good models consistently provide reliable input.
This allows the model to infer which are the reliable models in an unsupervised manner, i.e., without explicit supervision in the target language, and thereby make accurate inferences despite the substantial noise.

In the paper, we also consider a supervised setting, where a tiny annotated corpus is available in the target language.
We present two methods to use this data: 1) estimate reliability parameters of the Bayesian model, and  2) explicit model selection and fine-tuning of a low-resource supervised model, thus allowing for more accurate modelling of language specific parameters, such as character embeddings, shown to be important in previous work~\cite{xie2018neural}.


Experimenting on two NER corpora, one with as many as $41$ languages,
we show that single model transfer has highly variable performance, 
and uniform ensembling 
often substantially underperforms the single best model.
%
%
In contrast, our zero-shot approach does much better, exceeding the performance of the single best model, and our few-shot supervised models result in further gains.

\section{Approach}\label{sec:approach}
We frame the problem of multilingual transfer as follows.
We assume a collection of $H$ models, all trained in a high resource setting, denoted $M^h = \{M^h_i, i \in (1, H)\}$.
Each of these models are not well matched to our target data setting, for instance these may be trained on data from different domains, or on different languages, as we evaluate in our experiments, where we use cross-lingual embeddings for model transfer.
This is a problem of transfer learning, namely, how best we can use the $H$ models for best results in the target language.\footnote{We limit our attention to transfer in a `black-box' setting, that is, given predictive models, but not assuming access to their data, nor their implementation. This is the most flexible scenario, as it allows for application to settings with closed APIs, and private datasets. It does, however, preclude multitask learning, as the source models are assumed to be static.}

Simple approaches in this setting include a) choosing a single model $M \in M^h$, on the grounds of practicality, or the similarity between the model's native data condition and the target, and this model is used to label the target data; or b) allowing all models to `vote' in an classifier ensemble, such that the most frequent outcome is selected as the ensemble output. Unfortunately neither of these approaches are very accurate in a cross-lingual transfer setting,  as we show in \secref{sec:results}, where we show a fixed source language model (\texttt{en}) dramatically underperforms compared to oracle selection of source language, and the same is true for uniform voting.

Motivated by these findings, we propose novel methods for learning.
For the ``zero-shot'' setting where no labelled data is available in the target, we propose the \beauns method inspired by work in truth inference from crowd-sourced datasets or diverse classifiers (\S\ref{sec:bccmultiner}).
To handle the ``few-shot'' case \S\ref{sec:sagg} presents a rival supervised technique, \rare, based on using very limited annotations in the target language for model selection and classifier fine-tuning.


\subsection{Zero-Shot Transfer} 
\label{sec:bccmultiner}
\label{sec:bccmultiner:methods}

One way to improve the performance of the ensemble system is to select a subset of component models carefully, or more generally, learn a non-uniform weighting function.
Some models do much better than others, on their own, so it stands to reason that identifying these handful of models will give rise to better ensemble performance.
How might we proceed to learn the relative quality of models in the setting where no annotations are available in the target language?
This is a classic unsupervised inference problem, for which we propose a probabilistic graphical model, inspired by
\citet{kim2012bayesian}.

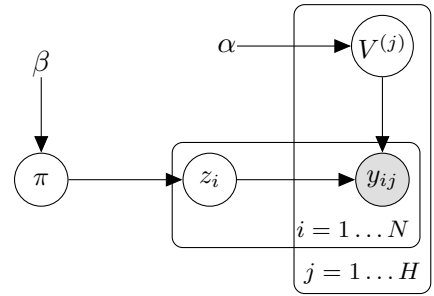
\begin{figure}
\centering
\begin{tikzpicture}
  \matrix[row sep=1cm, column sep=1.5cm] (DPC)
  {
    &
    &
    \node[latent] (v_j)  {$V^{(j)}$} ;
    \\
    \node[latent] (pi)   {$\pi$} ; &
    \node[latent] (z_i)  {$z_{i}$} ; &
    \node[obs]    (y_ij) {$y_{ij}$} ;
    \\
  };
  \node[const, above=of pi] (beta)  {$\beta$}; %
  \node[const, left=of v_j, xshift=-0.5cm] (alpha) {$\alpha$}; %

  \edge {beta}    {pi}  ; %
  \edge {pi}      {z_i} ; %
  \edge {alpha}   {v_j} ; %
  \edge {z_i,v_j} {y_ij} ; %

  \plate {plt_i} {
    (z_i)(y_ij)
  } {$i=1\dots N$} ;

  \plate {plt_j} {
    (v_j)(y_ij)(plt_i.south east)
  } {$j=1\dots H$} ;
\end{tikzpicture}
\caption{Plate diagram for the \bea model.}\label{fig:pgm_bcc}
\end{figure}

We develop a generative model, illustrated in \figref{fig:pgm_bcc}, of the transfer models' predictions, $y_{ij}$, where $i \in [1,N]$ is an instance (a token or an entity span), and $j \in [1,H]$ indexes a transfer model. 
The generative process assumes a `true' label, $z_i \in [1,K]$, which is corrupted by each transfer model, in producing the prediction, $y_{ij}$.
The corruption process is described by \mbox{$P(y_{ij}=l | z_i=k, V^{(j)}) = V^{(j)}_{kl}$},
where $V^{(j)} \in \mathcal{R}^{K \times K}$ is the confusion matrix specific to a transfer model.

To complete the story, the confusion matrices are drawn from vague row-wise independent Dirichlet priors, with a parameter $\alpha=1$,
and the true labels are governed by a Dirichlet prior, $\pi$, which is drawn from an uninformative Dirichlet distribution with a parameter $\beta=1$. This generative model is referred to as \bea.


Inference under the \bea model involves explaining the observed predictions $Y$ in the most efficient way.
Where several transfer models have identical predictions, $k$, on an instance, this can be explained by letting $z_i = k$,\footnote{Although there is no explicit breaking of the symmetry of the model, we initialise inference using the majority vote, which results in a bias towards this solution.} and the confusion matrices of those transfer models assigning high probability to $V_{kk}^{(j)}$.
Other, less reliable, transfer models will have divergent predictions, which are less likely to be in agreement, or else are heavily biased towards a particular class. Accordingly, the \bea model can better explain these predictions through label confusion, using the off-diagonal elements of the confusion matrix.
Aggregated over a corpus of instances, the \bea model can learn to differentiate between those reliable transfer models, with high $V^{(j)}_{kk}$ and those less reliable ones, with high $V^{(j)}_{kl}, ~l \ne k $.
This procedure applies per-label, and thus the `reliability' of a transfer model is with respect to a specific label, and may differ between classes.
This helps in the NER setting where many poor transfer models have excellent accuracy for the outside label, but considerably worse performance for entity labels.

For inference, we use mean-field variational Bayes~\cite{jordan1998learning}, which learns a variational distribution, $q(Z,V,\pi)$ to optimise the evidence lower bound (ELBO),
\begin{align*}
\log  P(Y|\alpha, \beta)  \nonumber
 \ge \mathbb{E}_{q(Z,V,\pi)} \log \frac{P(Y, Z, V, \pi|\alpha,\beta)}{q(Z,V,\pi)}
\end{align*}
assuming a fully factorised variational distribution, $q(Z,V,\pi)= q(Z)q(V)q(\pi)$.
This gives rise to an iterative learning algorithm with update rules:
\begin{subequations}\label{eq:rulepiandv}
\begin{align}
\mathbb{E}_q& \log\pi_k \label{eq:rulepi} \\
=& \psi\left(\beta+\sum_iq(z_i=k)\right) - \psi\left(K\beta+N\right) \nonumber\\
\mathbb{E}_q& \log V_{kl}^{(j)} \label{eq:rulev} \\
=& \psi\left(\alpha+\sum_iq(z_i=k)\mathbf{1}[y_{ij}=l]\right) \nonumber\\
&- \psi\left(K\alpha+\sum_iq(z_{i}=k)\right) \nonumber
\end{align}
\end{subequations}
\vspace{-2ex}
\begin{align}
\!\!\!\! q(z_{i}&=k)
 \propto \exp\left\{\mathbb{E}_q\log\pi_k+\sum_j\mathbb{E}_q\log V_{ky_{ij}}^{(j)}\right\}
\label{eq:rulez}
\end{align}
where $\psi$ is the digamma function, defined as the logarithmic derivative of the gamma function.
The sets of rules (\ref{eq:rulepiandv}) and (\ref{eq:rulez}) are applied alternately, to update the values of $\mathbb{E}_q\log\pi_k$, $\mathbb{E}_q\log V_{kl}^{(j)}$, and $q(z_{ij}=k)$ respectively.
This repeats until convergence, when the difference in the ELBO between two iterations is smaller than a threshold.

The final prediction of the model is based on $q(Z)$, using the maximum a posteriori label \mbox{$\hat{z}_{i} = \arg\max_{z} q(z_i = z)$.} This method is referred to as \beauns.

In our NER transfer task, classifiers are diverse in their F$1$ scores ranging from almost $0$ to around $80$, motivating \emph{spammer removal}~\cite{raykar2012eliminating} to filter out the worst of the transfer models.
We adopt a simple strategy that first estimates the confusion matrices for all transfer models on all labels, then ranks them based on their mean recall on different entity categories (elements on the diagonals of their confusion matrices), and then runs the \bea model again using only labels from the top $k$ transfer models only. We call this method \beaunstwice and its results are reported in \S\ref{sec:results}.

\subsubsection{Token versus Entity Granularity} 
\label{sec:uaggtok}
Our proposed aggregation method in \S\ref{sec:bccmultiner:methods} is based on an assumption that the true annotations are independent from each other, which simplifies the model but may generate undesired results.
That is, entities predicted by different transfer models could be mixed, resulting in labels inconsistent with the BIO scheme. Table~\ref{tab:token_entity_view} shows an example, where a sentence with $4$ words is annotated by $5$ transfer models with $4$ different predictions, among which at most one is correct as they overlap. However, the aggregated result in the token view is a mixture of two predictions, which is supported by no transfer models.

\begin{table}
\footnotesize
\begin{tabular}{@{}l|llll|lll@{}}
\toprule
        & $w_1$ & $w_2$ & $w_3$ & $w_4$ & $[1,4]$ & $[2,4]$ & $[3,4]$ \\ \midrule
$M_1^h$ & B-ORG & I-ORG & I-ORG & I-ORG & ORG     & O       & O       \\
$M_2^h$ & O     & B-ORG & I-ORG & I-ORG & O       & ORG     & O       \\
$M_3^h$ & O     & O     & B-ORG & I-ORG & O       & O       & ORG     \\
$M_4^h$ & O     & B-PER & I-PER & I-PER & O       & PER     & O       \\
$M_5^h$ & O     & B-PER & I-PER & I-PER & O       & PER     & O       \\ \midrule
Agg.    & O     & B-PER & I-ORG & I-ORG & O       & PER     & O       \\ \bottomrule
\end{tabular}
\caption{An example sentence with its aggregated labels in both token view and entity view. Aggregation in token view may generate results inconsistent with the BIO scheme.}
\label{tab:token_entity_view}
\end{table}

To deal with this problem, we consider aggregating the predictions in the entity view. As shown in Table~\ref{tab:token_entity_view}, we convert the predictions for tokens to predictions for ranges, aggregate labels for every range, and then resolve remaining conflicts.
A prediction is ignored if it conflicts with another one with higher probability.
By using this greedy strategy, we can solve the conflicts raised in entity-level aggregation. We use superscripts \tok and \ent to denote token-level and entity-level aggregations, i.e. \beauns[\tok] and \beauns[\ent].

\subsection{Few-Shot Transfer}
\label{sec:sagg}

Until now, we have assumed no access to annotations in the target language.
However, when some labelled text is available, how might this best be used?
In our experimental setting, we assume a modest set of 100 labelled sentences,
in keeping with a low-resource setting~\cite{garette2013twohours}.\footnote{\citet{garette2013twohours} showed that about $100$ sentences can be annotated with POS tags in two hours by non-native annotators.}
We propose two models \beasup and \rare in this setting.

\paragraph{Supervising \bea (\rbeasup)}\label{subsubsec:supervised_BEA}
One possibility is to use the labelled data to find the posterior for the parameters $V^{(j)}$ and $\pi$
of the Bayesian model described in \S\ref{sec:bccmultiner}. Let $n_k$ be the number of instances in the labelled data whose true label is $k$, and $n_{jkl}$ the number of instances whose true label is $k$ and classifier $j$ labels them as $l$. Then the quantities in Equation~\eqref{eq:rulepiandv} can be calculated as
\begin{align*}
\mathbb{E}\log\pi_k =& \psi(n_k) - \psi(N)\\
\mathbb{E}\log v_{jkl} =& \psi(n_{jkl}) - \psi\left(\sum_ln_{jkl}\right) \, .
\end{align*}
These are used in Equation~\eqref{eq:rulez} for inference on the test set.
We refer to this setting as \beasup.

\paragraph{Ranking and Retraining (\rsagg)}
We also propose an alternative way of exploiting the limited annotations,
\rare, which first \textbf{ra}nks the systems, and then uses the top ranked models' outputs
alongside the gold data to \textbf{re}train a model on the target language.
The motivation is that the above technique is agnostic to the input text, and
therefore is unable to exploit situations where regularities occur, such as
common words or character patterns that are indicative of specific class labels,
including names, titles, etc.
These signals are unlikely to be consistently captured by cross-lingual transfer.
Training a model on the target language with a character encoder component, can
distil the signal that are captured by the transfer models, while relating this
towards generalisable lexical and structural evidence in the target language.
This on its own will not be enough, as many tokens will be consistently
misclassified by most or all of the transfer models, and for this reason we also
perform model fine-tuning using the supervised data.

The ranking step in \rare proceeds by evaluating each of the $H$ transfer models
on the target gold set, to produce scores $s_h$ (using the \fone score). The scores are
then truncated to the top $k \le H$ values, such that $s_h = 0$ for those systems $h$
not ranked in the top $k$, and normalised $\omega_h = \frac{s_h}{\sum_{j=1}^k {s_j}}$.
The range of scores are quite wide, covering $0.00-0.81$ (see \figref{fig:dtsource}), and accordingly this
simple normalisation conveys a strong bias towards the top scoring transfer systems.

The next step is a distillation step, where a model is trained on a large unannotated
dataset in the target language, such that the model predictions match those of a
weighted mixture of transfer models, using $\vec{\omega}=(\omega_1,\dots,\omega_H)$ as the mixing weights.
This process is implemented as mini-batch scheduling, where the labels for
each mini-batch are randomly sampled from transfer model $h$ with probability $\omega_h$.\footnote{We show that uniform sampling with few source languages achieves worse performance.}
This is repeated over the course of several epochs of training.

Finally, the model is fine-tuned using the small supervised dataset, in order to correct for
phenomena that are not captured from model transfer, particularly character level information which
is not likely to transfer well for all but the most closely related languages.
Fine-tuning proceeds for a fixed number of epochs on the supervised dataset, to limit overtraining
of richly parameterise models on a tiny dataset.
Note that in all stages, the same supervised dataset is used, both in ranking and fine-tuning, and moreover,
we do not use a development set. 
This is not ideal, and generalisation performance would likely improve were we to use additional annotated data, however
our meagre use of data is designed for a low resource setting where labelled data is at a premium.

\section{Experiments}\label{sec:exp}
\subsection{Data}\label{sec:data}
Our primarily evaluation is over a subset of the \wikian NER corpus~\cite{pan2017cross}, using $41$ out of $282$ languages, where the langauges were chosen based on their overlap with multilingual word embedding resources from~\citet{conneau2017word}.\footnote{With ISO 639-1 codes: \zaf, \zar, \zbg, \zbn, \zbs, \zca, \zcs, \zda, \zde, \zel, \zen, \zes, \zet, \zfa, \zfi, \zfr, \zhe, \zhi, \zhr, \zhu, \zid, \zit, \zlt, \zlv, \zmk, \zms, \znl, \zno, \zpl, \zpt, \zro, \zru, \zsk, \zsl, \zsq, \zsv, \zta, \ztl, \ztr, \zuk and \zvi.}
The NER taggs are in IOB2 format comprising of \texttt{LOC}, \texttt{PER}, and \texttt{ORG}. The distribution of labels is highly skewed,
so we created balanced datasets, and partitioned into training, development, and test sets, details of which are in the Appendix.
For comparison with prior work, we also evaluate on the \texttt{CoNLL} 2002 and 2003 datasets~\cite{tjong2002conll,tjong2003conll}, which we discuss further in \secref{sec:results}.


For language-independent word embedding features we use \fasttext $300$ dimensional Wikipedia embeddings~\cite{bojanowski2017enriching},
and map them to the English embedding space using character-identical words as the seed for the Procrustes rotation method for learning bingual 
embedding spaces from \muse \cite{conneau2017word}.\footnote{
We also experimented with other bilingual embedding methods, including: supervised learning over bilingual dictionaries, which barely affected system performance;
and pure-unsupervised methods \cite{conneau2017word, artetxe2018acl}, which performed substantially worse. For this reason we use identical word type seeding, which is preferred as it imposes no additional supervision requirement.
}
Similar to \citet{xie2018neural} we don't
rely on a bilingual dictionary, so the method can be easily applied to other languages.

\begin{figure*}[t]
	\centering
	\begin{tikzpicture}[scale=1]
	\begin{axis}[
	xtick pos=left,
	    legend style={
		at={(0.75,-0.2)},
		anchor=north,
		legend columns=3
	},
	ytick pos=left,
	axis y line*=left,
	height=5cm,
	width=16cm,
	grid=major,
	enlarge x limits=false,
	axis x line*=left,
	ylabel near ticks,
	xlabel near ticks,
	xlabel=Target Language,
	ylabel=\fone,
	xmin=-1,xmax=29,
	ymin=1, ymax=90,
	xtick={0, 1, ..., 27},
	xticklabels={
	\rotatebox{90}{\zaf},\rotatebox{90}{\zar},\rotatebox{90}{\zbg},\rotatebox{90}{\zbn},\rotatebox{90}{\zbs},\rotatebox{90}{\zca},\rotatebox{90}{\zcs},\rotatebox{90}{\zde},\rotatebox{90}{\zel},\rotatebox{90}{\zet},\rotatebox{90}{\zfa},\rotatebox{90}{\zfi},\rotatebox{90}{\zfr},\rotatebox{90}{\zhe},\rotatebox{90}{\zhi},\rotatebox{90}{\zhr},\rotatebox{90}{\zhu},\rotatebox{90}{\zid},\rotatebox{90}{\zlt},\rotatebox{90}{\zlv},\rotatebox{90}{\zmk},\rotatebox{90}{\zms},\rotatebox{90}{\zro},\rotatebox{90}{\zta},\rotatebox{90}{\ztl},\rotatebox{90}{\ztr},\rotatebox{90}{\zuk},\rotatebox{90}{\zvi}},
	]
	
	\addplot[
	scatter/classes={a={red,mark=square*}},
	scatter,
	only marks,
	scatter src=explicit symbolic,
	nodes near coords*={\Label},
	visualization depends on={value \thisrow{label} \as \Label},
	visualization depends on={value \thisrow{anchor}\as\myanchor},
	every node near coord/.append style={anchor=\myanchor}
	]%
	table[meta=class, x=x, y=y]{
x l label class y anchor
0 af nl a 79 south
1 ar fa a 56 south
2 bg ru a 76 south
3 bn ar a 62 south
4 bs hr a 80 south
5 ca fr a 82 south
6 cs sk a 78 south
7 de nl a 70 south
8 el fr a 45 south
9 et fi a 76 south
10 fa de a 60 south
11 fi et a 78 south
12 fr it a 81 south
13 he es a 49 south
14 hi pt a 53 south
15 hr cs a 77 south
16 hu nl a 77 south
17 id it a 79 south
18 lt cs a 74 south
19 lv sk a 66 south
20 mk bg a 73 south
21 ms id a 74 south
22 ro es a 77 south
23 ta ar a 38 south
24 tl en a 74 south
25 tr nl a 71 south
26 uk ru a 60 south
27 vi ca a 54 south
	};

	\addplot[
	scatter/classes={en={blue},mv={black,mark=triangle*}},
	scatter,
	only marks,
	scatter src=explicit symbolic,
	]%
	table[meta=class, x=x, y=y]{
x lang y class

0 af 69 en
1 ar 18 en
2 bg 4 en
3 bn 44 en
4 bs 59 en
5 ca 76 en
6 cs 68 en
7 de 62 en
8 el 16 en
9 et 64 en
10 fa 8 en
11 fi 64 en
12 fr 71 en
13 he 25 en
14 hi 22 en
15 hr 63 en
16 hu 56 en
17 id 65 en
18 lt 61 en
19 lv 62 en
20 mk 2 en
21 ms 66 en
22 ro 58 en
23 ta 17 en
24 tl 74 en
25 tr 57 en
26 uk 4 en
27 vi 52 en

0 af 73 mv
1 ar 23 mv
2 bg 24 mv
3 bn 59 mv
4 bs 70 mv
5 ca 72 mv
6 cs 71 mv
7 de 69 mv
8 el 46 mv
9 et 73 mv
10 fa 28 mv
11 fi 77 mv
12 fr 70 mv
13 he 38 mv
14 hi 51 mv
15 hr 73 mv
16 hu 74 mv
17 id 54 mv
18 lt 75 mv
19 lv 66 mv
20 mk 23 mv
21 ms 68 mv
22 ro 64 mv
23 ta 44 mv
24 tl 53 mv
25 tr 65 mv
26 uk 18 mv
27 vi 45 mv

	};
	\legend{Top,En,MV}
	\end{axis}
	\end{tikzpicture}
	\caption{Best source language (\mysquare{red}) compared with \zen (\mycircle{blue}),
		and majority voting (\mytriangle{black}) over all source languages in terms of \fone performance in direct transfer shown for a subset of the $41$ target languages (x axis). Worst transfer score, not shown here, is about $0$. See \secref{sec:exp} for details of models and datasets. }
	\label{fig:dtsource}
\end{figure*}
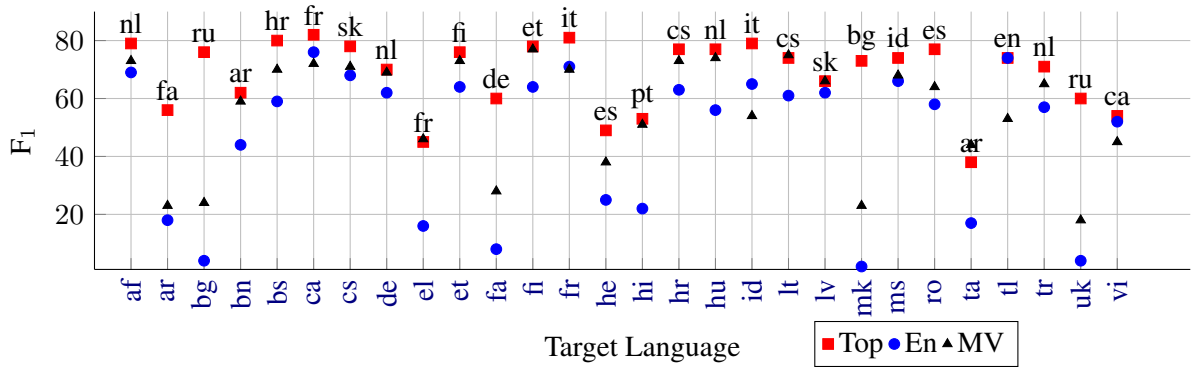

\subsection{Model Variations}

As the sequential tagger, we use a BiLSTM-CRF \cite{N16-1030}, which has been shown to result in state-of-the-art results in high resource settings~\cite{ma2016end,N16-1030}. This model includes both word embeddings (for which we used fixed cross-lingual embeddings) and character embeddings, to form a parameterised potential function in a linear chain conditional random field. With the exception of batch size and learning rate which were tuned (details in Appendix), we kept the architecture and the hyperparameters the same as the published code.\footnote{\url{https://github.com/guillaumegenthial/sequence_tagging}}

We trained models on all $41$ languages in both high-resource (\highsup) and naive supervised low-resource (\lowsup) settings, where \highsup pre-trained models were used for transfer in a leave-one-out setting, i.e., taking the predictions of 40 models into a single target language. The same BiLSTM-CRF is also used for \rare.

To avoid overfitting, we use early stopping based on a validation set for the \highsup, and \lowsup baselines.
For \ssagg, given that the model is already trained on noisy data, we stop fine-tuning after only $5$ iterations, chosen based on the performance for the first four languages.

We compare the supervised \textbf{\highsup} and \textbf{\lowsup} monolingual baselines with our proposed transfer models:

\vspace{-\topsep}
\begin{itemize}
	\setlength{\parskip}{0pt}
	\setlength{\itemsep}{0pt}
	\item[\textbf{\mv}] uniform ensemble, a.k.a.\@ ``majority vote'';
    \item[\textbf{\beaunstwice, \beauns}] unsupervised aggregation models, applied to entities or tokens (see \S\ref{sec:bccmultiner:methods});
	\item[\textbf{\beasup}] supervised estimation of \bea prior (\S\ref{sec:sagg});
	\item[\textbf{\rare, \rareuns}] supervised ranking and retraining model (\S\ref{sec:sagg}), and uniform ranking without fine-tuning, respectively; and
	\item[\textbf{\sbest}] selecting the best performing single transfer model, based on test performance.
\end{itemize}
\vspace{-\topsep}
We also compare with \bwet~\cite{xie2018neural} as state-of-the-art in unsupervised NER transfer. \bwet transfers
the source English training and development data to the target language using bilingual dictionary induction~\cite{conneau2017word},
and then uses a transformer architecture to compensate for missing sequential information. We used \bwet in both CoNLL, and \wikian datasets
by transferring from their corresponding source English data to the target language.\footnote{Because \bwet uses identical characters for
bilingual dictionary induction, we observed many English loan words in the target language mapped to the same word in the induced bilingual dictionaries.
Filtering such dictionary items might improve \bwet.}

\section{Results}\label{sec:results}

\begin{figure}[t]
	\centering
	\footnotesize

    \begin{tikzpicture}[scale=0.9]
	\begin{axis}[
	xtick pos=left,
	ytick pos=left,
	axis x line*=left,
	axis y line*=left,
	xlabel near ticks,
	ylabel near ticks,
	xlabel=\fone over $41$ langs.,
	ylabel=Annotation Requirement (\#sentences),
	grid=major,
	grid style={dashed,gray!30},
	ymin=0-70,ymax=1100+70,
	xmin=30-3, xmax=100+3,
	ytick={500,900,1000,1100},
	yticklabels={0, 100, 200, 5K+},
	]	
	\addplot[
	scatter/classes={a={bblue}, b={rred,mark=triangle*}, c={ggreen,mark=diamond*}, d={ppurple,mark=square*}},
	scatter,
	only marks,
	visualization depends on=\thisrow{std} \as \myshift,
	every node near coord/.append style = {anchor=west,shift={(axis direction cs:\myshift,0)}},
	scatter src=explicit symbolic,
	nodes near coords*={\Label},
	visualization depends on={value \thisrow{label} \as \Label},
	]%
	plot [error bars/.cd, x dir = both, x explicit]
	table[meta=class, x=x, y=y, x error=std]{
		y   x	  std  class  label
	   1100	89.21  2.82   d	  \highsup
	   1000	62.67  5.36   c	  \lowsup
		900	77.4  6.4   b	  \rare t10
		800	74.786775	 9.572119   b	  \beasup[\ent]t10
        700 72.757213	12.019092   b     \mvent t3
        600 70.864813	12.602649   b     \mvtok t3
        500 72.820458	11.508313   a     \beaunstwice[\ent]t10
		400	69.670004	12.594412   a	  \beauns[\ent]
		300 64.450713	13.744187   a     \beauns[\tok]
		200 62.7  5.3   a     \method{BWET}
		100 57.485519	26.122830   a     \mvent
		  0 56.744357	24.987837   a     \mvtok
	};
	\end{axis}
	\end{tikzpicture}
	\caption{
		The mean and standard deviation for the \fone score of the proposed unsupervised models (\beauns[\tok] and \beauns[\ent]), supervised models (\sagg and \beasup[\ent] t10) compared with state-of-the-art unsupervised model \bwet~\cite{xie2018neural}, high- and low-resource supervised models \highsup and \lowsup, and majority voting (\mvtok)
		in terms of entity level \fone over the $41$ languages ($40$ for \bwet) summarised from \tabref{tab:results}. The $x$ axis shows the annotation requirement of each model in the target language where ``200''
		means $100$ sentences each for training and development, and ``5K+"
		means using all the available annotation for training and development sets. Points with the same colour/shape have equal data requirement.}
	\label{fig:sumresults}
\end{figure}


We report the results for single source direct transfer, and then show that our proposed multilingual methods outperform
majority voting. Then we analyse the choice of source languages, and how it affects transfer.\footnote{For detailed results see \tabref{tab:results} in the Appendix.}
Finally we report results on
CoNLL NER datasets.
\paragraph{Direct Transfer}
The first research question we consider is the utility of direct transfer, and the simple majority vote ensembling method.
As shown in \figref{fig:dtsource}, using a single model for direct transfer (English: \texttt{en}) is often a terrible choice.
The oracle choice of source language model does much better, however it is not always a closely related language  (e.g., Italian: \texttt{it} does best for Indonesian: \texttt{id}, despite the target being closer to Malay: \texttt{ms}).
Note the collection of Cyrillic languages (\zbg, \zmk, \zuk) where the oracle is substantially better than the majority vote, which is likely due to script differences.
The role of script appears to be more important than language family, as seen for  Slavic languages where direct transfer works well between between pairs languages using the same alphabet 
(Cyrillic versus Latin), but much more poorly when there is an alphabet mismatch.\footnote{Detailed direct transfer results are shown in \figref{fig:transferheatmap} in the Appendix.}
The transfer relationship is not symmetric e.g., Persian: \texttt{fa} does best for Arabic: \texttt{ar}, but German: \texttt{de} does best for Persian.
\figref{fig:dtsource} also shows that ensemble voting is well below the oracle best language, which is likely to be a result of overall high error rates coupled with error correlation between models, and little can be gained from ensembling.

\paragraph{Multilingual Transfer}
We report the results for the proposed low-resource supervised models (\sagg and \beasup),
and unsupervised models (\beauns and \beaunstwice), summarised as an average over the $41$ languages in
 \figref{fig:sumresults} (see Appendix A for the full table of results).
The figure  compares against high- and low-resource supervised baselines (\highsup and \lowsup, respectively), and \method{BWET}.
The best performance is achieved with a high supervision (\highsup, $F_1=89.2$), while very limited supervision (\lowsup) results in a considerably lower \fone of  $62.1$.
The results for \mvtok show that uniform ensembling of multiple source models is even worse, by about $5$ points.

Unsupervised zero-shot learning dramatically improves upon \mvtok, and \beauns[\ent] outperforms  \beauns[\tok], showing the effectiveness of inference over entities rather than tokens.
It is clear that having access to limited annotation in the target language makes a substantial difference in \beasup[\ent] and \rare with \fone of $74.8$ and $77.4$, respectively.

\begin{figure}
	\centering
	\begin{tikzpicture}[scale=0.8]
	\begin{axis}[
	legend columns=3,
	legend style={at={(0.5,1.0)},anchor=south, font=\footnotesize},
	xmax=20,
	ymax=80,
    xlabel = \#source languages,
    ylabel = \fone,
	xtick pos=left,
	ytick pos=left,
	grid=both,
	grid style={dashed,gray!30},
	axis y line*=left,
	axis x line*=left,
	ylabel near ticks,
	xlabel near ticks
	]
	\plotfile{testdata.dat}
	\legend{\beaunstwice[\ent],\mvent,\beaunsent[\textit{, oracle}],\beasupent,\rare}
	\end{axis}
	\end{tikzpicture}
	\caption{The mean \fone performance of \mvent, \beasup[\ent], \beaunstwice[\ent], \beaunsent[\textit{, oracle}], and \rare over the $41$ languages by the number of source languages.}
	\label{fig:topk}
\end{figure}

Further analysis show that majority voting works reasonably well for Romance and Germanic languages, which are well represented in the dataset, but fails miserably compared to single best for Slavic languages (e.g.\ \zru, \zuk, \zbg)
where there are only a few related languages. For most of the isolated languages (\zar, \zfa, \zhe, \zvi, \zta), explicitly training
a model in \sagg outperforms \beasup[\ent],  showing that relying only on aggregation of annotated data has limitations, in that it cannot exploit character and structural features.


\paragraph{Choice of Source Languages}
An important question is how the other models, particularly the unsupervised variants, are affected by the number and choice of sources languages.
Figure~\ref{fig:topk} charts the performance of \mv, \bea, and \rare against the number of source models, comparing the use of ideal or realistic selection methods to attempt to find the best source models. \mvent, \beasup[\ent], and \rare use a small labeled dataset to rank the source models. \beaunsent[\textit{, oracle}] has the access to the perfect ranking of source models based on their real \fone on the test set. \beaunstwice is completely unsupervised in that it uses its own estimates to rank all source models.

\mv doesn't show any benefit with more than $3$ source models.\footnote{The sawtooth pattern arises from the increased numbers of ties (broken randomly) with even numbers of inputs.}
In contrast, \bea and \rare continue to improve with up to $10$ languages. We show that
\bea in two realistic scenarios (unsupervised: \beaunstwice[\ent], and supervised: \beasup[\ent]) is highly effective at discriminating between good and bad source models, and thus filtering out the bad models gives the best results. The \beaunstwice[\ent] curve shows the effect of filtering using purely unsupervised signal, which has a positive, albeit mild effect on performance. In \beaunsent[\textit{, oracle}] although the source model ranking is perfect, it narrowly outperforms \bea. Note also that neither of the \bea curves show evidence of the sawtooth pattern, i.e., they largely benefit from more inputs, irrespective of their parity. Finally, adding supervision in the target language in \rare further improves upon the unsupervised models.

\paragraph{CoNLL Dataset}
Finally, we apply our model to the CoNLL-02/03 datasets, to benchmark our technique against related work.
This corpus is much less rich than \wikian used above, as it includes only four languages (en, de, nl, es), and furthermore, the languages are closely related and share the same script.
Results in \tabref{tab:conll} show that our methods are competitive with benchmark methods,
and, moreover, the use of $100$ annotated sentences
in the target language (\raress{l}) gives good improvements over unsupervised models.\footnote{For German because of its capitalisation pattern, we lowercase all the source and target data, and also remove German as a source model for other languages.}
Results also show that \mv does very well, especially \mvent, and its performance is comparable to \bea's. Note that there are only 3 source models and none of them is clearly bad, so \bea estimates that they are similarly reliable which results in little difference in terms of performance between \bea and \mv.
\begin{table}
	\centering
	\begin{tabular}{l *{4}{ S[table-format=2.1,round-mode=places,round-precision=1] }}
		\toprule
		lang. 							& {de} 		& {es} 		& {nl} 		& {en}\\
		\midrule
		\citet{tackstrom2012cluster}\textsuperscript{$p$}	& 40.40 	& 59.30 	& 58.40 	& \multicolumn{1}{l}{---} \\
		\citet{joel2013wikiner}\textsuperscript{$w$} 		& 55.80 	& 61.00 	& 64.00 	& 61.30 \\
		\citet{tsai2016wiki}\textsuperscript{$w$} 			& 48.12 	& 60.55 	& 61.60 	& \multicolumn{1}{l}{---} 	\\
		\citet{ni2017weakly}\textsuperscript{$w,p,d$} 		& 58.50 	& 65.10 	& 65.40 	& \multicolumn{1}{l}{---} 	\\
		\citet{mayhew2017cheap}\textsuperscript{$w,d$}  	& 59.11 	& 65.95 	& 66.50 	& \multicolumn{1}{l}{---} 	\\
		\citet{xie2018neural}\textsuperscript{$0$} 			& 57.76 	& 72.37 	& 70.40 	& \multicolumn{1}{l}{---}	\\
		\midrule
		our work \\
        $\mv^{\tok,\ 0}$ & 57.38 & 66.41 & 71.01 & 62.14 \\
        $\mv^{\ent,\ 0}$ & 57.67 & 69.03 & 70.34 & 64.64 \\
        \beauns[\tok,\ 0] & 58.18 & 64.72 & 70.14 & 61.24 \\
        \beauns[\ent,\ 0] & 57.76 & 63.37 & 70.30 & 64.81 \\
		\rareunsss{0} 						& 59.14		& 71.75	 	& 67.59  	& 67.46 \\
		\raress{l}						 	& 63.99  	& 72.49  	& 72.45  	& 70.04 \\
		\midrule
		\highsup 						& 79.11 	& 85.69 	& 87.11 	& 89.50 	\\
		\bottomrule
	\end{tabular}
	
	\caption{The performance of \rare and \bea in terms of phrase-based \fone on CoNLL NER datasets compared with state-of-the-art benchmark methods. Resource requirements are indicated with superscripts, $p$: parallel corpus, $w$: Wikipedia, $d$: dictionary, $l$: $100$ NER annotation, $0$: no extra resources.}
	\label{tab:conll}
\end{table}




\section{Related Work}\label{sec:related}
Two main approaches for cross-lingual transfer are representation and annotation projection.
Representation projection learns a model in a high-resource source language using
representations that are cross-linguistically transferable, and then directly applies the model to
data in the target language.
This can include the use of cross-lingual word clusters~\cite{tackstrom2012cluster} and
word embeddings~\cite{ammar2016many,ni2017weakly}, multitask learning with a closely related high-resource language (e.g.\ Spanish for Galician)~\cite{I17-2016}, or
bridging the source and target languages through phonemic transcription~\cite{bharadwaj2016phono} or Wikification~\cite{tsai2016wiki}.
In annotation projection, the annotations of tokens in a source sentence are projected to their aligned tokens in the target language through a parallel corpus. Annotation projection has been applied to
POS tagging~\cite{yarowsky2001inducing,das2011unsup,long2014mulipos,fang2016distant},
NER~\cite{zitouni2008mention,ehrmann2011multiner,agerri2018parallel}, and parsing~\cite{hwa2005bootstrapping,ma2014unsupervised,rasooli2015parsing,rasooli2015density}. The Bible, Europarl, and recently the Watchtower has been used as parallel corpora, which are limited in genre, size, and language coverage, motivating the use of Wikipedia to create weak annotation for multilingual tasks such as NER~\cite{joel2013wikiner}.
Recent advances in (un)supervised bilingual dictionary induction~\cite{gouws2015biembs,long2016crosslingual,conneau2017word,artetxe2018acl,schuster2019cross} have enabled
cross-lingual alignment with bilingual dictionaries~\cite{mayhew2017cheap,xie2018neural}. Most annotation projection methods with few exceptions~\cite{tackstrom2012nudging,plank2018distant} use only one language (often English) as the source language. In multi-source language setting, majority voting is often used to aggregate noisy annotations (e.g.\ ~\citet{plank2018distant}). \citet{fang2016distant} show the importance of modelling the annotation biases that the source language(s) might project to the target language.

\paragraph{Transfer from multiple source languages:}
Previous work has shown the improvements of multi-source transfer in NER~\cite{tackstrom2012nudging,moe2017active,enghoff2018parallel},
POS tagging~\cite{snyder2009morelang,plank2018distant}, and parsing~\cite{ammar2016many} compared to single source transfer, however, multi-source transfer might be noisy as a result of divergence in script, phonology, morphology, syntax, and semantics between the source languages, and the target language. To capture such differences, various methods have been proposed: latent variable models~\cite{snyder2009morelang}, majority voting~\cite{plank2018distant}, utilising typological features~\cite{ammar2016many}, or explicitly learning annotation bias~\cite{moe2017transfer}. Our work is also related to knowledge distillation from multiple source models applied in parsing~\cite{kuncoro2016distillation} and machine translation\cite{kim2016distillation,johnson2017zeroshot}. In this work, we use truth inference to model the transfer annotation bias from diverse source models.

Finally, our work is related to truth inference from crowd-sourced annotations~\citep{whitehill2009whose,welinder2010multidimensional},
and most importantly from diverse classifiers~\citep{kim2012bayesian,ratner2017snorkel}.
\citet{nguyen2017aggregating} propose a hidden Markov model for aggregating crowdsourced sequence labels,
but only learn per-class accuracies for workers instead of full confusion matrices 
in order to address the data sparsity problem in crowdsourcing.



\section{Conclusion}\label{sec:conclusion}

Cross-lingual transfer does not work out of the box, especially when using large numbers of source languages, and distantly related target languages.
In an NER setting using a collection of $41$ languages, we showed that simple methods such as uniform ensembling do not work well.
We  proposed two new multilingual transfer models (\rare and \bea), based on unsupervised transfer, or a supervised transfer setting with a small $100$ sentence labelled dataset in the target language. We also compare our results with \bwet~\cite{xie2018neural}, a state-of-the-art unsupervised single source (English) transfer model, and showed that multilingual transfer outperforms
it, however, our work is orthogonal to their work in that if training data from multiple source models is created, \rare and \bea can still combine them, and outperform majority voting.
Our unsupervised method,  \beauns, provides a fast and simple way of annotating data in the target language, which is capable of reasoning under noisy annotations, and outperforms several competitive baselines, including the majority voting ensemble, a low-resource supervised baseline, and the oracle single best transfer model.
We show that light supervision improves performance further, and that our second approach, \rare, based on ranking transfer models and then retraining on the target language, results in further and more consistent performance improvements.

\section*{Acknowledgments}
This work was supported by a Facebook Research Award and the Defense Advanced Research Projects Agency Information Innovation Office (I2O), under the Low Resource
Languages for Emergent Incidents (LORELEI)
program issued by DARPA/I2O under Contract
No. HR0011-15-C-0114. The views expressed are
those of the author and do not reflect the official
policy or position of the Department of Defense
or the U.S. Government.

\bibliography{acl2019}
\bibliographystyle{acl_natbib}

\clearpage
\appendix

\section{Appendices}
\label{sec:appendix}
\subsection{Hyperparameters}
We tuned the batch size and the learning rate using development sets in four languages,\footnote{Afrikaans, Arabic, Bulgarian and Bengali.} and then fixed these hyperparameters for all other languages in each model.
The batch size was $1$ sentence in low-resource scenarios (in baseline \lowsup and fine-tuning of \rare), and to $100$ sentences, in high-resource settings (\highsup and the pretraining phase of \rare). The learning rate was set to $0.001$ and $0.01$ for the high-resource and low-resource baseline models, respectively,
and to $0.005$, $0.0005$ for the pretraining and fine-tuning phase of \ssagg based on development results for the four languages.
For CoNLL datasets, we had to decrease the batch size of the pre-training phase from $100$ to $20$ (because of GPU memory issues).

\subsection{Cross-lingual Word Embeddings}
We experimented with Wiki and CommonCrawl monolingual embeddings from \fasttext \cite{bojanowski2017enriching}. Each of the $41$ languages is mapped to English embedding space using three methods from \muse: 1) supervised with bilingual dictionaries; 2) seeding using identical character sequences; and 3) unsupervised training using adversarial learning~\cite{conneau2017word}. The cross-lingual mappings are evaluated by precision at $k=1$. The resulting cross-lingual embeddings are then used in NER direct transfer in a leave-one-out setting for the $41$ languages ($41 \times 40$ transfers), and we report the mean \fone in \tabref{tab:embeddings}. CommonCrawl doesn't perform well in bilingual induction despite having larger text corpora, and underperforms in direct transfer NER. It is also evident that using identical character strings instead of a bilingual dictionary as the seed for learning a supervised bilingual mapping barely affects the performance. 
This finding also applies to few-shot learning over larger ensembles: running \rare over 40 source  languages achieves an average \fone of $77.9$ when using embeddings trained with a dictionary, versus $76.9$ using string identity instead. 
For this reason we have used the string identity method in the paper (e.g., \tabref{tab:results}), providing greater portability to language pairs without a bilingual dictionary.
Experiments with unsupervised mappings performed substantially worse than supervised methods, and so we didn't explore these further.

\begin{table}[t]
	\centering
	\begin{tabular}{llcc}
		\toprule
		& & Transl. Acc. & Dir.Transf. \fone\\
		\midrule 
		\multirow{2}{*}{Unsup} & crawl & 34 & 26\\ & wiki & 24 & 21 \\
		\multirow{2}{*}{IdentChar} & crawl & 43 & 37\\ & wiki & \textbf{53} & \textbf{44} \\
		\multirow{2}{*}{Sup} & crawl & 50 & 39\\ & wiki & \textbf{54} & \textbf{45} \\
		\bottomrule
	\end{tabular}
	\caption{
		The effect of the choice of monolingual word embeddings (Common Crawl and Wikipedia), and their cross-lingual mapping on NER direct transfer. Word translation accuracy, and direct transfer NER \fone are averaged over $40$ languages.}
	\label{tab:embeddings}
\end{table}

\subsection{Direct Transfer Results}
\begin{figure*}[t]
	\centering
	\includegraphics[width=\textwidth]{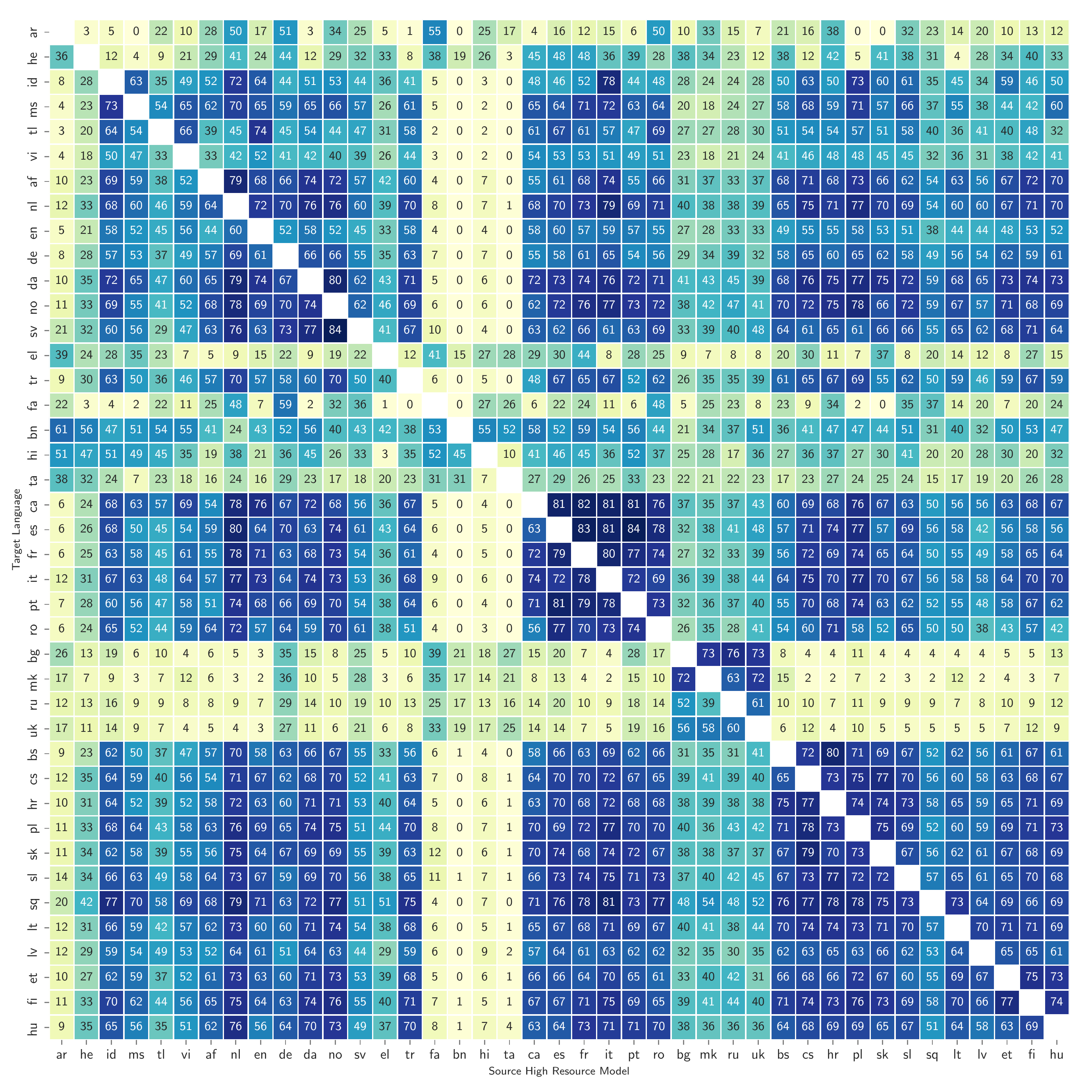}
	\caption{
		The direct transfer performance of a source NER model trained in a high-resource setting applied on the other $40$ target languages, and evaluated in terms of phrase-level \fone. The languages are roughly sorted by language family. Slavic languages in Cyrillic
		script are from \zbg to \zuk, and those in Latin script are from \zbs to \zsl.}
	\label{fig:transferheatmap}
\end{figure*}

In \figref{fig:transferheatmap} the performance of an NER model trained in a high-resource setting on a source language applied on the other $40$ target languages (leave-one-out) is shown.
An interesting finding is that symmetry does not always hold (e.g. \zid vs. \zms or \zfa vs. \zar).

\subsection{Detailed Low-resource Results}
The result of applying baselines, proposed models and their variations, and unsupervised transfer model
of \citet{xie2018neural} are shown in \tabref{tab:results}.
\begin{table*}[tbp]
\footnotesize
	\centering
    \sisetup{table-format = 2.1, table-number-alignment = right}
	\begin{tabular}{@{}lSS|S|SS|SSSS|SSSSSS|S@{}}
\toprule
&&& \multicolumn{7}{c|}{\textsf{Supervised}} & \multicolumn{6}{c|}{\textsf{Unsupervised}} & \\
\cmidrule(l{0.3em}r{0.3em}){5-10} \cmidrule(l{0.3em}r{0.3em}){11-16}
	& \rot{\method{\#train(k)}} & \rot{\method{\#test(k)}} & \rot{\method{BiDic.P@1}}& \rot{\highsup} & \rot{\lowsup} & \rot{\rare t1} & \rot{\rare t10} & \rot{\rare all} & \rot{\beasup[\ent] t10} & \rot{\rareuns} & \rot{\bwet} & \rot{\beaunstwice[\ent] t10} & \rot{\beauns[\ent]} & \rot{\beauns[\tok]} & \rot{\mvtok} & \rot{\sbest} \\
\midrule
af       & 5    & 1    & 36   & 84   & 59   & 73   & 79   & 79   & 80   & 76   & 64   & 79   & 79   & 74   & 75   & 80   \\
ar       & 20   & 10   & 46   & 88   & 64   & 71   & 74   & 74   & 65   & 26   & 19   & 54   & 45   & 54   & 12   & 56   \\
bg       & 20   & 10   & 55   & 90   & 61   & 80   & 81   & 81   & 81   & 5    & 51   & 81   & 65   & 54   & 4    & 76   \\
bn       & 10   & 1    & 1    & 95   & 70   & 68   & 74   & 74   & 69   & 65   & 36   & 67   & 66   & 60   & 56   & 63   \\
bs       & 15   & 1    & 30   & 92   & 63   & 80   & 79   & 80   & 78   & 76   & 52   & 80   & 78   & 77   & 69   & 82   \\
ca       & 20   & 10   & 70   & 91   & 62   & 82   & 86   & 84   & 86   & 80   & 62   & 85   & 80   & 79   & 72   & 83   \\
cs       & 20   & 10   & 64   & 90   & 62   & 77   & 78   & 75   & 78   & 73   & 59   & 77   & 75   & 72   & 71   & 78   \\
da       & 20   & 10   & 68   & 90   & 62   & 77   & 81   & 81   & 82   & 79   & 68   & 83   & 82   & 79   & 78   & 80   \\
de       & 20   & 10   & 73   & 86   & 58   & 73   & 74   & 73   & 72   & 69   & 63   & 72   & 71   & 64   & 68   & 70   \\
el       & 20   & 10   & 55   & 89   & 61   & 67   & 67   & 67   & 54   & 13   & 45   & 49   & 43   & 34   & 13   & 45   \\
en       & 20   & 10   & \multicolumn{1}{l|}{---}   & 81   & 47   & 64   & 65   & 64   & 65   & 58   & \multicolumn{1}{l}{---}   & 63   & 61   & 57   & 56   & 61   \\
es       & 20   & 10   & 83   & 90   & 63   & 83   & 84   & 84   & 85   & 76   & 62   & 85   & 81   & 76   & 73   & 84   \\
et       & 15   & 10   & 41   & 90   & 64   & 73   & 77   & 77   & 78   & 72   & 58   & 78   & 78   & 71   & 73   & 75   \\
fa       & 20   & 10   & 33   & 93   & 74   & 78   & 81   & 79   & 69   & 30   & 16   & 65   & 50   & 52   & 15   & 60   \\
fi       & 20   & 10   & 58   & 89   & 67   & 78   & 80   & 80   & 81   & 76   & 68   & 81   & 80   & 69   & 77   & 78   \\
fr       & 20   & 10   & 82   & 88   & 57   & 81   & 81   & 80   & 84   & 75   & 59   & 83   & 79   & 73   & 71   & 80   \\
he       & 20   & 10   & 52   & 85   & 53   & 61   & 61   & 60   & 55   & 40   & 26   & 54   & 54   & 46   & 34   & 50   \\
hi       & 5    & 1    & 29   & 85   & 68   & 64   & 74   & 73   & 68   & 48   & 27   & 64   & 61   & 58   & 35   & 54   \\
hr       & 20   & 10   & 48   & 89   & 61   & 74   & 79   & 78   & 80   & 76   & 49   & 80   & 79   & 77   & 73   & 78   \\
hu       & 20   & 10   & 64   & 90   & 59   & 75   & 79   & 78   & 80   & 71   & 55   & 79   & 79   & 69   & 73   & 76   \\
id       & 20   & 10   & 68   & 91   & 67   & 82   & 83   & 81   & 75   & 59   & 62   & 73   & 67   & 61   & 62   & 79   \\
it       & 20   & 10   & 77   & 89   & 60   & 80   & 81   & 80   & 82   & 75   & 59   & 81   & 78   & 76   & 72   & 79   \\
lt       & 10   & 10   & 26   & 86   & 62   & 72   & 79   & 80   & 79   & 76   & 48   & 80   & 80   & 75   & 77   & 74   \\
lv       & 10   & 10   & 31   & 91   & 68   & 70   & 75   & 75   & 69   & 68   & 40   & 69   & 69   & 67   & 65   & 66   \\
mk       & 10   & 1    & 50   & 91   & 67   & 79   & 82   & 81   & 80   & 4    & 38   & 79   & 66   & 48   & 3    & 75   \\
ms       & 20   & 1    & 48   & 91   & 66   & 78   & 80   & 78   & 74   & 69   & 62   & 68   & 67   & 63   & 68   & 74   \\
nl       & 20   & 10   & 76   & 89   & 59   & 78   & 80   & 80   & 81   & 77   & 63   & 82   & 81   & 78   & 76   & 79   \\
no       & 20   & 10   & 67   & 90   & 65   & 79   & 82   & 81   & 83   & 79   & 59   & 83   & 83   & 77   & 79   & 79   \\
pl       & 20   & 10   & 66   & 89   & 61   & 76   & 79   & 78   & 81   & 73   & 63   & 82   & 80   & 77   & 76   & 78   \\
pt       & 20   & 10   & 80   & 90   & 59   & 79   & 81   & 80   & 82   & 77   & 65   & 82   & 77   & 74   & 70   & 82   \\
ro       & 20   & 10   & 67   & 92   & 66   & 80   & 82   & 82   & 80   & 76   & 46   & 78   & 76   & 74   & 67   & 77   \\
ru       & 20   & 10   & 59   & 86   & 53   & 73   & 71   & 71   & 56   & 10   & 38   & 53   & 40   & 36   & 11   & 61   \\
sk       & 20   & 10   & 52   & 91   & 62   & 76   & 79   & 79   & 80   & 74   & 50   & 79   & 76   & 76   & 71   & 79   \\
sl       & 15   & 10   & 47   & 92   & 64   & 76   & 80   & 80   & 79   & 76   & 58   & 79   & 78   & 76   & 73   & 78   \\
sq       & 5    & 1    & 37   & 88   & 69   & 79   & 84   & 84   & 83   & 82   & 59   & 83   & 84   & 76   & 79   & 79   \\
sv       & 20   & 10   & 61   & 93   & 69   & 83   & 83   & 84   & 82   & 77   & 60   & 79   & 80   & 69   & 76   & 84   \\
ta       & 15   & 1    & 7    & 84   & 54   & 44   & 53   & 53   & 46   & 35   & 12   & 39   & 42   & 25   & 29   & 38   \\
tl       & 10   & 1    & 20   & 93   & 66   & 75   & 82   & 80   & 78   & 65   & 60   & 62   & 60   & 57   & 52   & 76   \\
tr       & 20   & 10   & 61   & 90   & 61   & 75   & 77   & 77   & 77   & 70   & 53   & 77   & 76   & 67   & 67   & 71   \\
uk       & 20   & 10   & 45   & 89   & 60   & 70   & 78   & 79   & 70   & 5    & 35   & 64   & 58   & 49   & 6    & 60   \\
vi       & 20   & 10   & 54   & 88   & 55   & 64   & 72   & 72   & 61   & 58   & 53   & 56   & 55   & 48   & 47   & 56   \\ \midrule
$\mu$    & \multicolumn{1}{l}{---} & \multicolumn{1}{l|}{---} & \multicolumn{1}{l|}{---} & 89.2 & 62.1 & 74.3 & 77.4 & 76.9 & 74.8 & 60.2 & 50.5 & 72.8 & 69.7 & 64.5 & 56.7 & 71.6 \\
$\sigma$ & \multicolumn{1}{l}{---} & \multicolumn{1}{l|}{---} & \multicolumn{1}{l|}{---} & 2.8  & 5.2  & 7.3  & 6.4  & 6.4  & 9.6  & 24.1 & 14.7 & 11.5 & 12.6 & 13.7 & 25   & 11.5 \\ \bottomrule
\end{tabular}
\caption{The size of training and test sets (development set size equals test set size) in thousand sentences, and the precision at $1$ for Bilingual dictionaries induced from mapping languages to the English embedding space (using identical characters) is shown (\method{BiDic.P@1}). \fone scores on the test set, comparing baseline supervised models (\highsup, \lowsup),
multilingual transfer from top $k$ source languages (\rare, $5$ runs, $k=1,10, 40$),
an unsupervised \rare with uniform expertise and no fine-tuning (\rareuns), and
aggregation methods: majority voting (\mvtok), \beauns[\tok] and \beauns[\ent] (Bayesian aggregation in token- and entity-level), and the oracle single best annotation (\sbest).
We also compare with \bwet \cite{xie2018neural}, an unsupervised transfer model with state-of-the-art on CoNLL NER datasets. The mean and standard deviation over all $41$ languages, $\mu, \sigma$, are also reported.}\label{tab:results}
\end{table*}


\end{document}